\documentclass[10pt,twocolumn,letterpaper]{article}

\usepackage{cvpr}
\usepackage{times}
\usepackage{epsfig}
\usepackage{graphicx}
\usepackage{amsmath}
\usepackage{amssymb}
\usepackage{array}
\usepackage{tabulary}
\newcolumntype{K}[1]{>{\centering\arraybackslash}m{#1}}
\usepackage{subcaption}
\usepackage{flushend}
\usepackage{cite}

\usepackage[breaklinks=true,bookmarks=false, backref=page]{hyperref}

\cvprfinalcopy % *** Uncomment this line for the final submission

\def\cvprPaperID{5050} % *** Enter the CVPR Paper ID here

\begin{document}

\title{VarifocalNet: An IoU-aware Dense Object Detector}
\author{Haoyang Zhang$^{1}$, Ying Wang$^{2}$, Feras Dayoub$^{1}$, Niko S\"underhauf$^{1}$ \\
$^{1}$Australian Centre for Robotic Vision, Queensland University of Technology \\
$^{2}$University of Queensland\\
{\tt\small \{h202.zhang, feras.dayoub, niko.suenderhauf\}@qut.edu.au, ying.wang@uq.edu.au}
}

\maketitle

\begin{abstract}
Accurately ranking the vast number of candidate detections is crucial for dense object detectors to achieve high performance. Prior work uses the classification score or a combination of classification and predicted localization scores to rank candidates. However, neither option results in a reliable ranking, thus degrading detection performance. In this paper, we propose to learn an Iou-aware Classification Score (\textbf{IACS}) as a joint representation of object presence confidence and localization accuracy. We show that dense object detectors can achieve a more accurate ranking of candidate detections based on the IACS. We design a new loss function, named \textbf{Varifocal Loss}, to train a dense object detector to predict the IACS, and propose a new star-shaped bounding box feature representation for IACS prediction and bounding box refinement. Combining these two new components and a bounding box refinement branch, we build an IoU-aware dense object detector based on the FCOS+ATSS architecture, that we call \textbf{VarifocalNet} or \textbf{VFNet} for short. Extensive experiments on MS COCO show that our VFNet consistently surpasses the strong baseline by $\sim$2.0 AP with different backbones. Our best model VFNet-X-1200 with Res2Net-101-DCN achieves a single-model single-scale AP of \textbf{55.1} on COCO test-dev, which is state-of-the-art among various object detectors. Code is available at: https://github.com/hyz-xmaster/VarifocalNet.

\end{abstract}

\section{Introduction}
Modern object detectors, regardless of being a two-stage method~\cite{RCNN, fastRCNN, fasterRCNN, maskRCNN} or a one-stage method~\cite{YOLO, YOLOv3, SSD, retinaNet, FCOS}, usually first generate a redundant set of bounding boxes with classification scores and then deploy non-maximum suppression (NMS) to remove duplicated bounding boxes on the same object. Generally, the classification score is used to rank the bounding box in NMS~\cite{RCNN, fastRCNN, fasterRCNN, maskRCNN, retinaNet}.
However, this harms the detection performance, because the classification score is not always a good estimate of the bounding box localization accuracy~\cite{IoUNet} and accurately localized detections with low classification scores may be mistakenly removed in NMS. 

\begin{figure}[t]
	\begin{center}
		\includegraphics[width=1.0\columnwidth]{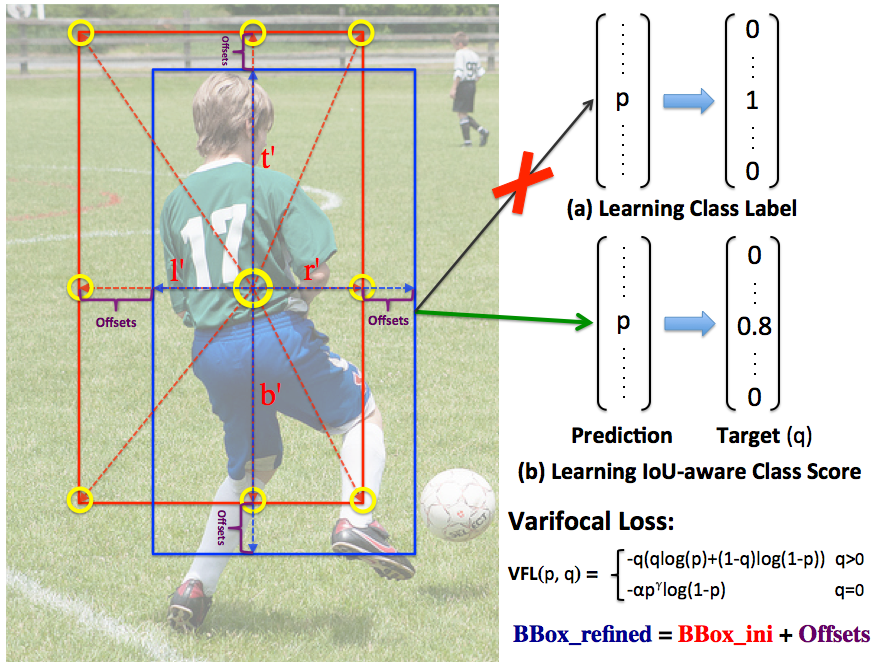}
	\end{center}
	\vspace{-5mm}
	\caption{An illustration of our method. Instead of learning to predict the class label (a) for a bounding box, we learn the IoU-aware classification score (\textbf{IACS}) as its detection score which merges the object presence confidence and localization accuracy (b). We propose a \textbf{varifocal loss} for training a dense object detector to predict the IACS, and a star-shaped bounding box feature representation (the features at nine yellow sampling points) for IACS prediction. With the new representation, we refine the initially regressed box (in red) into a more accurate one (in blue).}
	\label{fig:overview}
	\vspace{-6mm}
\end{figure}
To solve the problem, existing dense object detectors predict either an additional IoU score~\cite{IoURetinaNet} or a centerness score~\cite{FCOS} as the localization accuracy estimation, and multiply them by the classification score to rank detections in NMS. These methods can alleviate the misalignment problem between the classification score and the object localization accuracy. However, they are sub-optimal because multiplying the two imperfect predictions may lead to a worse rank basis and we show in Section~\ref{sec:motivation} that the upper bound of the performance achieved by such methods is limited. Besides, adding an extra network branch to predict the localization score is not an elegant solution and incurs additional computation burden.

To overcome these shortcomings, we naturally would like to ask:
\textit{Instead of predicting an additional localization accuracy score, can we merge it into the classification score?} That is, predict a localization-aware or IoU-aware classification score (\textbf{IACS}) that simultaneously represents the presence of a certain object class and the localization accuracy of a generated bounding box.

In this paper, we answer the above question and make the following contributions. 
\textbf{(1)} We show that accurately ranking candidate detections is critical for high performing dense object detectors, and IACS achieves a better ranking than other methods
(Section~\ref{sec:motivation}).
\textbf{(2)} We propose a new \textbf{Varifocal Loss} for training dense object detectors to regress the IACS. \textbf{(3)} We design a new  star-shaped bounding box feature representation for computing the IACS and refining the bounding box. \textbf{(4)} We develop a new dense object detector based on the FCOS~\cite{FCOS}+ATSS~\cite{ATSS} and the proposed components, named \textbf{VarifocalNet} or \textbf{VFNet} for short, to exploit the advantage of the IACS. An illustration of our method is shown in Figure~\ref{fig:overview}.

The Varifocal Loss, inspired by the focal loss~\cite{retinaNet}, is a dynamically scaled binary cross entropy loss. However, it supervises the dense object detector to regress continuous IACSs, and more distinctively it adopts an \textbf{asymmetrical} training example weighting method. It down-weights only negative examples for addressing the class imbalance problem during training, and yet up-weights high-quality positive examples for generating prime detections. This focuses the training on high-quality positive examples, which is important to achieve a high detection performance.

The star-shaped bounding box feature representation uses the features at nine fixed sampling points (yellow circles in Figure~\ref{fig:overview}) to represent a bounding box with the deformable convolution~\cite{DCN, DCNv2}. Compared to the point feature used in most existing dense object detectors~\cite{SSD, retinaNet, FCOS, foveaBox}, this feature representation can capture the geometry of the bounding box and its nearby contextual information, which is essential for predicting an accurate IACS. It also enables us to effectively refine the initially generated coarse bounding box without losing efficiency.

To verify the effectiveness of our proposed modules, we build the VFNet based on the FCOS+ATSS and evaluate it on COCO benchmark~\cite{COCO}.  Experiments show that our VFNet consistently exceeds the strong baseline by $\sim$2.0 AP with different backbones, and our best model VFNet-X-1200 with Res2Net-101-DCN reaches a single-model single-scale \textbf{55.1} AP on COCO \texttt{test-dev}, surpassing previously published best single-model single-scale results.

\section{Related Work}
\paragraph{Object Detection.}
With the development of object detection, currently popular object detectors can be categorized by whether they use anchor boxes or not. While popular two-stage methods~\cite{fasterRCNN, maskRCNN} and multi-stage methods~\cite{cascadeRCNN} usually employ anchors to generate object proposals for downstream classification and regression, anchor-based one-stage methods~\cite{YOLOv3, SSD, retinaNet, freeAnchor, ATSS, guidedAnchoring} directly classify and refine anchor boxes without object proposal generation. 

More recently, anchor-free detectors have attracted substantial attention due to their novelty and simplicity. One kind of them formulates the object detection problem as a key-point or a semantic-point detection problem, including CornerNet~\cite{cornerNet}, CenterNet\cite{centerNet}, ExtremeNet~\cite{extremeNet}, ObjectsAsPoints~\cite{objectsAsPoints} and RepPoints~\cite{repPoints}. Another type of anchor-free detectors are similar to anchor-based one-stage methods, but they remove the usage of anchor boxes. Instead, they classify each point on the feature pyramids~\cite{FPN} into foreground classes or background, and directly predict the distances from the foreground point to the four sides of the ground-truth bounding box, to produce the detection. Popular methods include DenseBox~\cite{denseBox}, FASF~\cite{FSAF}, FoveaBox~\cite{foveaBox}, FCOS~\cite{FCOS}, and SPAD~\cite{SAPD}. We build our VFNet based on the ATSS~\cite{ATSS} version of FCOS due to its simplicity, high efficiency and excellent performance.

\vspace{-4mm}
\paragraph{Detection Ranking Measures.}
In addition to the classification score, other detection ranking measures have been proposed.
IoU-Net~\cite{IoUNet} adopts an additional network to predict the IoU and uses it to rank bounding boxes in NMS, but it still selects the classification score as the final detection score. Fitness NMS~\cite{fitnessNMS}, IoU-aware RetinaNet~\cite{IoURetinaNet} and~\cite{rankProposals} are similar to IoU-Net in essence, except that they multiply the predicted IoU or IoU-based ranking scores and the classification score as the ranking basis. Instead of predicting the IoU-based score, FCOS~\cite{FCOS} predicts centerness scores to suppress the low-quality detections.

By contrast, we predict only the IACS as the ranking score. This avoids the overhead of an additional network and the possible worse ranking basis resulting from multiplying the imperfect localization and classification scores.

\vspace{-4mm}
\paragraph{Encoding the Bounding Box.}
Extracting discriminative features to represent a bounding box is important for downstream classification and regression in object detection. 
In two-stage and multi-stage methods, RoI Pooling~\cite{fastRCNN, fasterRCNN} or RoIAlign~\cite{maskRCNN} are widely employed to extract bounding box features. But applying them in dense object detectors is time-consuming. 
Instead, one-stage detectors generally use point features as the bounding box descriptor~\cite{SSD, retinaNet, FCOS}, due to the efficiency consideration. However, these local features fail to capture the geometry of the bounding box and essential contextual information.

Alternatively, HSD~\cite{HSD} and RepPoints~\cite{repPoints} extract features at learned semantic points with the deformable convolution to encode a bounding box. However, learning to localize semantic points is challenging due to the lack of strong supervision, and the prediction of semantic points also aggravates the computation burden.

In comparison, our proposed star-shaped bounding box representation uses the features at nine fixed sampling points to describe a bounding box. It is simple, efficient, and yet able to capture the geometry of the bounding box and spatial context cues around it.

\vspace{-4.5mm}
\paragraph{Generalized Focal Loss.}
The most similar work to ours is a concurrent work, Generalized Focal Loss (GFL)~\cite{GFL}. The GFL extends the focal loss~\cite{retinaNet} to a continuous version and trains a detector to predict a joint representation of localization quality and classification. 

We emphasize first that our varifocal loss is a distinct function from the GFL. It weights positive and negative examples asymmetrically, whereas the GFL deals with them equally, and experiment results show that our varifocal loss performs better than the GFL. Moreover, we propose a star-shaped bounding box feature representation to facilitate the prediction of IACS, and further improve the object localization accuracy through a bounding box refinement step, which are not considered in the GFL.

\section{Motivation}\label{sec:motivation}
\begin{figure}[t]
	\begin{center}
		\includegraphics[width=0.9\columnwidth]{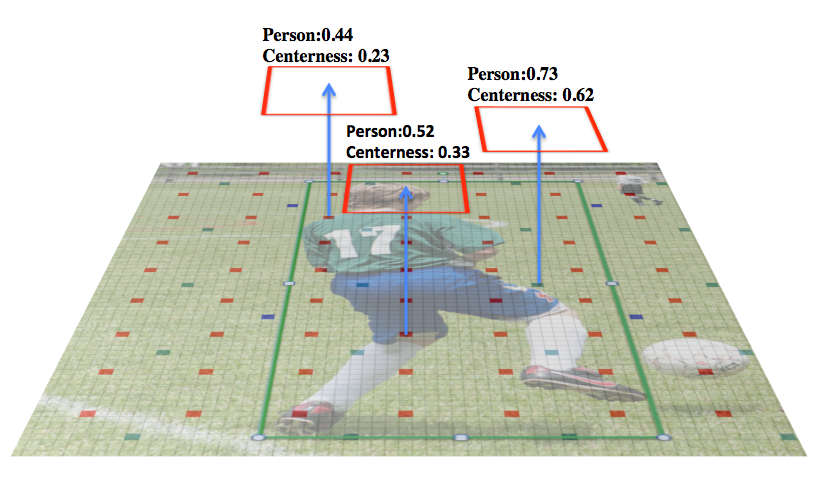}
	\end{center}
	\vspace{-7.5mm}
	\caption{An example of the output from the FCOS head which includes a classification score, a bounding box and a centerness score.}
	\label{fig:fcos}
\vspace{-2mm}
\end{figure}

\setlength{\tabcolsep}{1pt}
\begin{table}[t]
	\begin{center}
		\begin{tabular}{c | c c c c | c c | c c | c c }
			\hline%\noalign{\smallskip}
			 & \multicolumn{10}{c}{FCOS+ATSS} \\
			\hline
			w/ctr &  & \checkmark  & \checkmark & \checkmark &  & \checkmark &  & \checkmark &  & \checkmark\\
			gt\_ctr &  &  & \checkmark  &  & & & & & &  \\
			gt\_ctr\_iou &  &  &    & \checkmark & & & & & &   \\
            gt\_bbox &  &  & & & \checkmark & \checkmark & & & &\\
            gt\_cls &  &  &  &  & &  &\checkmark & \checkmark &  &\\
			gt\_cls\_iou &  &  &  & &  & & & &\checkmark & \checkmark\\ 
			\hline%\noalign{\smallskip}
            AP  & 38.5 & 39.2 & 41.1 & \multicolumn{1}{c}{43.5}  & 56.1 & \multicolumn{1}{c}{56.3} & 43.1 & \multicolumn{1}{c}{58.1} & \textbf{74.7} & 67.4\\
			\hline
		\end{tabular}
	\end{center}
	\vspace{-5mm}
	\caption{Performance of the FCOS+ATSS on COCO \texttt{val2017} with different oracle predictions. W/ctr means using the centerness score in inference. Please see the text for the meaning of other abbreviations.}
	\label{table:fcos}
\vspace{-6mm}
\end{table}
\setlength{\tabcolsep}{1.4pt}

In this section, we investigate the performance upper bound of a popular anchor-free dense object detector, FCOS~\cite{FCOS}, identify its main performance hindrance and show the importance of producing the IoU-aware classification score as the ranking criterion.

FCOS is built on FPN~\cite{FPN} and its detection head has three branches. One predicts the classification score for each point on the feature map, one regresses the distances from the point to the four sides of a bounding box, and another predicts the centerness score which is multiplied by the classification score to rank the bounding box in NMS. Figure~\ref{fig:fcos} shows an example of the output from the FCOS head. In this paper, we actually study the ATSS version of FCOS (FCOS+ATSS) in which the Adaptive Training Sample Selection (ATSS) mechanism~\cite{ATSS} is used to define foreground and background points on the feature pyramids during training. We refer the reader to~\cite{ATSS} for more details.

To investigate the performance upper bound of the FCOS+ATSS (trained on COCO \texttt{train2017}~\cite{COCO}), we alternately replace the predicted classification score, the distance offsets and the centerness score with corresponding ground-truth values for foreground points \textbf{before} NMS, and evaluate the detection performance in terms of AP~\cite{COCO} on COCO \texttt{val2017}. For the classification score vector, we implement two options, that is, replacing its element at the ground-truth label position with a value of 1.0 or the IoU between the predicted bounding box and the ground-truth one (termed as gt\_IoU). We also consider replacing the centerness score with the gt\_IoU in addition to its true value.

The results are shown in Table~\ref{table:fcos}. We can see that the original FCOS+ATSS achieves 39.2 AP. When using the ground-truth centerness score (gt\_ctr) in inference, unexpectedly, only about 2.0 AP is increased. Similarly, replacing the predicted centerness score with the gt\_IoU (gt\_ctr\_iou) only achieves 43.5 AP. This indicates that using the product of either the predicted centerness score or the IoU score and the classification score to rank detections is certainly unable to bring significant performance gain.

By contrast, the FCOS+ATSS with ground-truth bounding boxes (gt\_bbox) achieves 56.1 AP even without centerness score (no w/ctr) in inference. But if setting the classification score as 1.0 at the ground-truth label position (gt\_cls), whether or not to use the centerness score becomes important (43.1 AP vs 58.1 AP). Because the centerness score can differentiate accurate and inaccurate boxes to some extent.

The most \textbf{surprising result} is the one obtained by replacing the classification score of the ground-truth class with the gt\_IoU (gt\_cls\_iou). Without the centerness score, this case achieves an impressive \textbf{74.7} AP which is significantly higher than other cases. This in fact reveals that there already exist accurately localized bounding boxes in the large candidate pool for most objects. The key to achieving an excellent detection performance is to accurately select those high-quality detections from the pool, and these results show that replacing the classification score of the ground-truth class with the gt\_IoU is the most promising selection measure. We refer to the element of such a score vector as the \textbf{IoU-aware Classification Score (IACS)}.

\section{VarifocalNet}

\begin{figure*}[th]
	\begin{center}
		\includegraphics[width=1.0\linewidth]{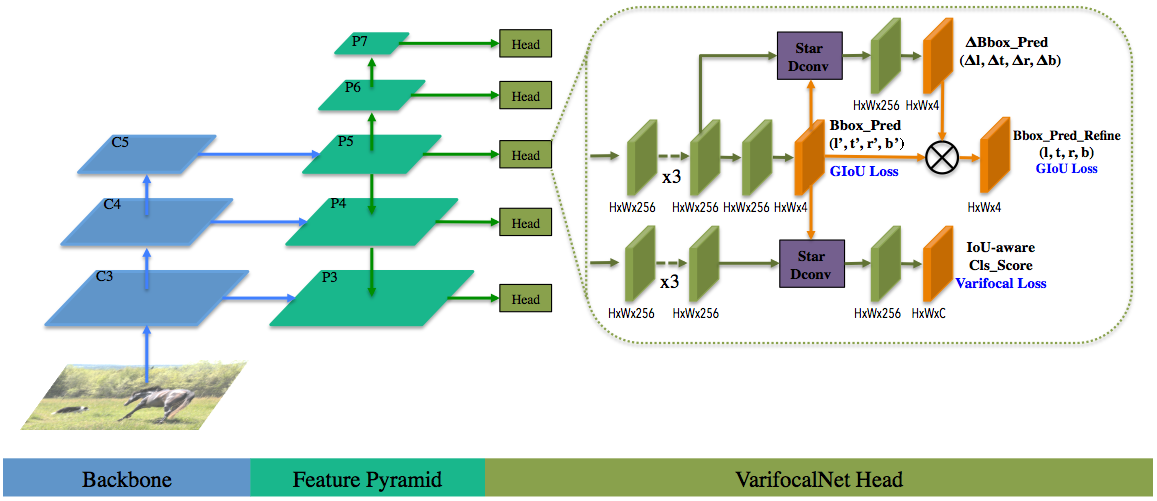}
	\end{center}
	\vspace{-5mm}
	\caption{The network architecture of our VFNet. The VFNet is built on the FPN (P3-P7). Its head consists of two subnetworks, one for regressing the initial bounding box and refining it, and the other for predicting the IoU-aware classification score, based on a star-shaped bounding box feature representation (Star Dconv). \texttt{H}$\times$\texttt{W} denotes the size of the feature map. }
	\label{fig:VFNet}
\vspace{-3mm}
\end{figure*}
Based on the discovery above, we propose to learn the IoU-aware classification score (IACS) to rank detections. To this end, we build a new dense object detector, coined as VarifocalNet or VFNet, based on the FCOS+ATSS with the centerness branch removed. Compared with the FCOS+ATSS, it has three new components: the varifcoal loss, the star-shaped bounding box feature representation and the bounding box refinement.

\subsection{IACS -- IoU-Aware Classification Score} 
We define the IACS as a scalar element of a classification score vector, in which the value at the ground-truth class label position is the IoU between the predicted bounding box and its ground truth, and 0 at other positions.

\subsection{Varifocal Loss}
We design the novel~\textit{Varifocal Loss} for training a dense object detector to predict the IACS. Since it is inspired by~\textit{Focal Loss}~\cite{retinaNet}, we first briefly review the focal loss. 

Focal loss is designed to address the extreme imbalance problem between foreground and background classes during the training of dense object detectors. It is defined as:
\begin{equation}
\mathrm{FL(p, y)} =  \begin{cases} \mathrm{- \alpha (1-p)^{\gamma}log(p)} & \mathrm{if \quad y  =  1} \\ \mathrm{- (1 - \alpha) p^{\gamma}log(1-p)}  & \mathrm{otherwise,}  \end{cases} 
\label{eq:FL}
\end{equation}
where $y \in \{ \pm 1\}$ specifies the ground-truth class and $p \in [0, 1]$ is the predicted probability for the foreground class.
As shown in Equation~\ref{eq:FL}, the modulating factor ($(1-p)^{\gamma}$ for the foreground class and $p^{\gamma}$ for the background class) can reduce the loss contribution from easy examples and relatively increases the importance of mis-classified examples. Thus, the focal loss prevents the vast number of easy negatives from overwhelming the detector during training and focuses the detector on a sparse set of hard examples.

We borrow the example weighting idea from the focal loss to address the class imbalance problem when training a dense object detector to regress the continuous IACS. However, unlike the focal loss that deals with positives and negatives equally, we treat them asymmetrically. Our varifocal loss is also based on the binary cross entropy loss and is defined as:
\begin{equation}
\mathrm{VFL(p, q)} =  \begin{cases} \mathrm{-q(qlog(p)+(1-q)log(1-p))} & \mathrm{q  >  0} \\ \mathrm{- \alpha  p^{ \gamma }log(1-p)}  & \mathrm{q = 0,}  \end{cases} 
\label{eq:VFL}
\end{equation}
where $p$ is the predicted IACS and $q$ is the target score. For a foreground point, $q$ for its ground-truth class is set as the IoU between the generated bounding box and its ground truth (gt\_IoU) and 0 otherwise, whereas for a background point, the target $q$ for all classes is 0. See Figure~\ref{fig:overview}.

As Equation~\ref{eq:VFL} shows, the varifocal loss only reduces the loss contribution from negative examples (q=0) by scaling their losses with a factor of $ p^{ \gamma }$ and does not down-weight positive examples (q$>$0) in the same way.
This is because positive examples are extremely rare compared with negatives and we should keep their precious learning signals. On the other hand, inspired by PISA~\cite{PISA} and~\cite{IoUBalancedLoss}, we weight the positive example with the training target $q$. If a positive example has a high gt\_IoU, its contribution to the loss will thus be relatively big. This focuses the training on those high-quality positive examples which are more important for achieving a higher AP than those low-quality ones.

To balance the losses between positive examples and negative examples, we add an adjustable scaling factor $\alpha$ to the negative loss term.

\subsection{Star-Shaped Box Feature Representation}
We design a star-shaped bounding box feature representation for IACS prediction. It uses the features at nine fixed sampling points (yellow circles in Figure~\ref{fig:overview}) to represent a bounding box with the deformable convolution~\cite{DCN, DCNv2}. This new representation can capture the geometry of a bounding box and its nearby contextual information, which is essential for encoding the misalignment between the predicted bounding box and the ground-truth one.

Specifically, given a sampling location (x, y) on the image plane (or a projecting point on the feature map), we first regress an initial bounding box from it with 3x3 convolution. Following the FCOS, this bounding box is encoded by a 4D vector (l', t', r', b') which means the distance from the location (x, y) to the left, top, right and bottom side of the bounding box respectively. With this distance vector, we heuristically select nine sampling points at: (x, y), (x-l', y), (x, y-t'), (x+r', y), (x, y+b'), (x-l', y-t'), (x+l', y-t'), (x-l', y+b') and (x+r', y+b'), and then map them onto the feature map. Their relative offsets to the projecting point of (x, y) serve as the offsets to the deformable convolution~\cite{DCN, DCNv2} and then features at these nine projecting points are convolved by the deformable convolution to represent a bounding box. Since these points are manually selected without additional prediction burden, our new representation is computation efficient.

\subsection{Bounding Box Refinement}
We further improve the object localization accuracy through a bounding box refinement step. Bounding box refinement is a common technique in object detection~\cite{cascadeRCNN, refineDet}, however, it is not widely adopted in dense object detectors due to the lack of an efficient and discriminative object descriptor. With our new star representation, we can now adopt it in dense object detectors without losing efficiency. 

We model the bounding box refinement as a residual learning problem. For an initially regressed bounding box (l', t', r', b'), we first extract the star-shaped representation to encode it. Then, based on the representation, we learn four distance scaling factors ($\Delta$l, $\Delta$t, $\Delta$r, $\Delta$b) to scale the initial distance vector, so that the refined bounding box that is represented by (l, t, r, b) = ($\Delta$l$\times$l', $\Delta$t$\times$t', $\Delta$r$\times$r', $\Delta$b$\times$b') is closer to the ground truth.

\subsection{VarifocalNet}
Attaching the above three components to the FCOS network architecture and removing the original centerness branch, we get the \textit{VarifocalNet}. 

Figure~\ref{fig:VFNet} illustrates the network architecture of the VFNet. The backbone and FPN network parts of the VFNet are the same as the FCOS. The difference lies in the head structure. The VFNet head consists of two subnetworks. The localization subnet performs bounding box regression and subsequent refinement. It takes as input the feature map from each level of the FPN and first applies \textbf{three} 3x3 conv layers with ReLU activations. This produces a feature map with 256 channels. One branch of the localization subnet convolves the feature map again and then outputs a 4D distance vector (l', t', r', b') per spatial location which represents the initial bounding box. Given the initial box and the feature map, the other branch applies a star-shaped deformable convolution to the nine feature sampling points and produces the distance scaling factor ($\Delta$l, $\Delta$t, $\Delta$r, $\Delta$b) which is multiplied by the initial distance vector to generate the refined bounding box (l, t, r, b). 

The other subnet aims to predict the IACS. It has the similar structure to the localization subnet (the refinement branch) except that it outputs a vector of C (the class number) elements per spatial location, where each element represents jointly the object presence confidence and localization accuracy.

\subsection{Loss Function and Inference}
\paragraph{Loss Function.}
The training of our VFNet is supervised by the loss function:
\begin{equation}
\begin{split}
\mathrm{Loss} & = \mathrm{\frac{1}{N_{pos}}\sum\limits_{i} \sum\limits_{c} VFL(p_{c,i},\: q_{c,i})} \\
& + \mathrm{\frac{\lambda_{0}}{N_{pos}}\sum\limits_{i}q_{c^{*},i}L_{bbox}(bbox'_{i}, \: bbox_{i}^{*})}  \\
& + \mathrm{\frac{\lambda_{1}}{N_{pos}}\sum\limits_{i}q_{c^{*},i}L_{bbox}(bbox_{i}, \: bbox_{i}^{*})} 
\end{split}
\label{eq:loss}
\end{equation}
where $p_{c,i}$ and $q_{c,i}$ denote the predicted and target IACS respectively for the class c at the location i on each level feature map of FPN. $L_{bbox}$ is the GIoU loss~\cite{GIoU}, and $bbox'_{i}$, $bbox_{i}$ and $bbox_{i}^{*}$ represent the initial, refined and ground-truth bounding box respectively. We weight the $L_{bbox}$ with the training target $q_{c^{*},i}$, which is the gt\_IoU for foreground points and 0 otherwise, following the FCOS. $\lambda_{0}$ and $\lambda_{1}$ are the balance weights for $L_{bbox}$ and are empirically set as 1.5 and 2.0 respectively in this paper. $N_{pos}$ is the number of foreground points and is used to normalize the total loss. As mentioned in Section~\ref{sec:motivation}, we employ the ATSS~\cite{ATSS} to define foreground and background points during training. 

\vspace{-4mm}
\paragraph{Inference.}
The inference of the VFNet is straightforward. It involves simply forwarding an input image through the network and a NMS post-processing step for removing redundant detections.

\section{Experiments}
\paragraph{Dataset and Evaluation Metrics.}
We evaluate the VFNet on the challenging MS COCO 2017 benchmark~\cite{COCO}. Following the common practice~\cite{fasterRCNN, FCOS, retinaNet, ATSS}, we train detectors on the \texttt{train2017} split, report ablation results on the \texttt{val2017} split and compare with other detectors on the \texttt{test-dev} split by uploading the results to the evaluation server. We adopt the standard COCO-style Average Precision (AP) as the evaluation metrics. 

\vspace{-4mm}
\paragraph{Implementation and Training Details.} 
We implement the VFNet with MMDetection~\cite{mmdetection}. Unless specified, we adopt the default hyper-parameters used in MMDetection. The initial learning rate is set as 0.01 and we employ the linear warming up policy~\cite{1hour} to start the training where the warm-up ratio is set as 0.1. We use 8 V100 GPUs for training with a total batch size of 16 (2 images per GPU) in both ablation studies and performance comparison.

For ablation studies on the \texttt{val2017}, the ResNet-50~\cite{ResNet} is used as the backbone network and 1x training schedule (12 epochs)~\cite{mmdetection} is adopted. Input images are resized to a maximum scale of 1333$\times$800, without changing the aspect ratio. Only random horizontal image flipping is used for data augmentation. 

For performance comparison with the state-of-the-art on the \texttt{test-dev}, we train the VFNet with different backbone networks, including those ones with deformable convolution layers~\cite{DCN, DCNv2} (denoted as DCN) inserted. When DCN is used in the backbone, we also insert it into the last layers before the star deformable convolution in the VFNet head.  2x (24 epochs) training scheme and multi-scale training (MSTrain) are adopted, where a maximum image scale for each iteration is randomly selected from a scale range. In fact, we apply two image scale ranges in experiments. For fair comparison with the baseline, we use the scale range 1333$\times$[640:800]; out of curiosity, we also experiment with a wider scale range 1333$\times$[480:960]. Note that even MSTrain is employed, we keep the maximum image scale to 1333$\times$800 in inference, although a bigger scale performs slightly better (about 0.4 AP gain with 1333$\times$900 scale).

\vspace{-4mm}
\paragraph{Inference Details.} 
In inference, we forward the input image which is resized to a maximum scale of 1333$\times$800 through the network and obtain estimated bounding boxes with corresponding IACSs. We first filter out those bounding boxes with $p_{max} \leq$  0.05 and select at most 1k top-scoring detections per FPN level. Then, the selected detections are merged and redundant detections are removed by NMS with a threshold of 0.6 to yield the final results.

\begin{table}[t]
    \begin{center}
        \begin{tabular}{K{2.5em} K{2.5em} K{5.5em} | K{3em} K{3em} K{3em} }
            \hline
            $\gamma$ & $\alpha$ & q weighting & AP &  AP$_{50}$ &  AP$_{75}$\\
            \hline
             1.0 & 0.50 & \checkmark & 41.2  & 59.2  & 44.7  \\ 
             1.5 & 0.75 & \checkmark & 41.5  & 59.7  & 45.1  \\
             2.0 & 0.75 & \checkmark & \textbf{41.6}  & 59.5  & 45.0  \\
             2.0 & 0.75 &            & 41.2  & 59.1  & 44.4  \\
             2.5 & 1.25 & \checkmark & 41.5  & 59.4  & 45.2  \\ 
             3.0 & 1.00 & \checkmark & 41.3  & 59.0  & 44.7  \\ 
            \hline
        \end{tabular}
    \end{center}
    \vspace{-5mm}
\caption{Peformance of the VFNet when changing the hyper-parameters ($\alpha$, $\gamma$) of the varifocal loss. q weighting means weighting the loss of the positive example with the learning target q.  }
\label{table:VFL}
\vspace{-0mm}
\end{table}

\begin{table}[t]
    \begin{center}
        \begin{tabular}{ K{2.5em} K{2.5em} K{4.5em} | K{3em} K{3em} K{3em} }
            \hline
             VFL  & Star Dconv & BBox Refinement & AP &  AP$_{50}$ &  AP$_{75}$\\
            \hline
                        &   &  & 39.0 & 57.7 & 41.8 \\
             \checkmark &   &  & 40.1 & 58.5 & 43.4 \\ 
             \checkmark & \checkmark &  & 40.7  & 59.0  & 44.0  \\ 
             \checkmark & \checkmark & \checkmark & \textbf{41.6}  & 59.5  & 45.0  \\ 
            \hline
            \multicolumn{3}{c|}{FCOS+ATSS} & 39.2 & 57.3 & 42.4 \\
            \hline
        \end{tabular}
    \end{center}
    \vspace{-5mm}
\caption{Individual contribution of the components in our method. The first row represents the results of the raw VFNet trained with the focal loss~\cite{retinaNet}.}
\label{table:ablation}
\vspace{-2mm}
\end{table}

\begin{table*}[tpb]
    \begin{center}
        \begin{tabular}{ K{11.0em} | K{8.0em} |K{2.5em} | K{4.2em} K{4.2em} K{4.2em} | K{4.2em} K{4.2em} K{4.2em} } 
            \hline
             Method  & Backbone & FPS & AP &  AP$_{50}$ &  AP$_{75}$ & AP$_{S}$ & AP$_{M}$ & AP$_{L}$ \\
            \hline
             Anchor-based multi-stage: &    &  & &  &  &  &\\ 
             Faster R-CNN~\cite{fasterRCNN} & X-101 & & 40.3 & 62.7 & 44.0 & 24.4 & 43.7 & 49.8  \\
             Libra R-CNN~\cite{libraRCNN} & R-101 & & 41.1 & 62.1 & 44.7 & 23.4 & 43.7 & 52.5  \\
             Mask R-CNN~\cite{maskRCNN} & X-101 &  & 41.4 & 63.4 & 45.2 & 24.5 & 44.9 & 51.8  \\
             R-FCN~\cite{RFCN} & R-101 &  & 41.4 & 63.4 & 45.2 & 24.5 & 44.9 & 51.8  \\
             TridentNet~\cite{TridentNet} & R-101  &  & 42.7 & 63.6 & 46.5 & 23.9 & 46.6 & 56.6  \\
             Cascade R-CNN~\cite{cascadeRCNN} & R-101 &  & 42.8 & 62.1 & 46.3 & 23.7 & 45.5 & 55.2  \\
             SNIP~\cite{SNIP} & R-101 &  & 43.4 & 65.5 & 48.4 & 27.2 & 46.5 & 54.9  \\

             \hline
             Anchor-based one-stage: &    &  &  &  &  &\\
             SSD512~\cite{SSD} & R-101 &  & 31.2 & 50.4 & 33.3 & 10.2 & 34.5 & 49.8 \\
             YOLOv3~\cite{YOLOv3} & DarkNet-53 &  & 33.0 & 57.9 & 34.4 & 18.3 & 35.4 & 41.9  \\
             DSSD513~\cite{DSSD} & R-101 &  & 33.2 & 53.3 & 35.2 & 13.0 & 35.4 & 51.1 \\
             RefineDet~\cite{refineDet} & R-101 &  & 36.4 & 57.5 & 39.5 & 16.6 & 39.9 & 51.4 \\
             RetinaNet~\cite{retinaNet} & R-101 &  & 39.1 & 59.1 & 42.3 & 21.8 & 42.7 & 50.2 \\
             FreeAnchor~\cite{freeAnchor} & R-101 &  & 43.1 & 62.2 & 46.4 & 24.5 & 46.1 & 54.8 \\
             GFL~\cite{GFL} & R-101-DCN &  & 47.3 & 66.3 & 51.4 & 28.0 & 51.1 & 59.2  \\
             GFL~\cite{GFL} & X-101-32x4d-DCN &  & 48.2 & 67.4 & 52.6 & 29.2 & 51.7 & 60.2  \\
             
            EfficientDet-D6~\cite{EfficientDet} & B6 & 5.3$^{\dag}$ & 51.7 & 71.2 & 56.0 & 34.1 & 55.2 & 64.1  \\
            EfficientDet-D7~\cite{EfficientDet} & B6 & 3.8$^{\dag}$ & 52.2 & 71.4 & 56.3 & 34.8 & 55.5 & 64.6  \\
             
             \hline
             Anchor-free key-point: &    &  &  &  &  &\\ 
             ExtremeNet~\cite{extremeNet} & Hourglass-104 &  & 40.2 & 55.5 & 43.2 & 20.4 & 43.2 & 53.1  \\
             CornerNet~\cite{cornerNet} & Hourglass-104 &  & 40.5 & 56.5 & 43.1 & 19.4 & 42.7 & 53.9  \\
             Grid R-CNN~\cite{gridRCNN} & X-101 &  & 43.2 & 63.0 & 46.6 & 25.1 & 46.5 & 55.2  \\
             CenterNet~\cite{cornerNet} & Hourglass-104 &  & 44.9 & 62.4 & 48.1 & 25.6 & 47.4 & 57.4  \\
             RepPoints~\cite{repPoints} & R-101-DCN &  & 45.0 & 66.1 & 49.0 & 26.6 & 48.6 & 57.5  \\
             
             \hline
             Anchor-free one-stage: &    &  &  &  &  &\\
             FoveaBox~\cite{foveaBox} & X-101 &  & 42.1 & 61.9 & 45.2 & 24.9 & 46.8 & 55.6  \\
             FSAF~\cite{FSAF} & X-101-64x4d &  & 42.9 & 63.8 & 46.3 & 26.6 & 46.2 & 52.7  \\
             FCOS~\cite{FCOS} & R-101 &   & 43.0 & 61.7 & 46.3 & 26.0 & 46.8 & 55.0  \\
             SAPD~\cite{SAPD} & R-101 &   & 43.5 & 63.6 & 46.5 & 24.9 & 46.8 & 54.6  \\
             SAPD~\cite{SAPD} & R-101-DCN &  & 46.0 & 65.9 & 49.6 & 26.3 & 49.2 & 59.6  \\

             \hline
             Baseline: &    &  &  &  &  &\\
             ATSS~\cite{ATSS} & R-101 & 17.5 & 43.6 & 62.1 & 47.4 & 26.1 & 47.0 & 53.6 \\
             ATSS~\cite{ATSS} & X-101-64x4d & 8.9 & 45.6 & 64.6 & 49.7 & 28.5 & 48.9 & 55.6 \\
             ATSS~\cite{ATSS} & R-101-DCN & 13.7 & 46.3 & 64.7 & 50.4 & 27.7 & 49.8 & 58.4 \\
             ATSS~\cite{ATSS} & X-101-64x4d-DCN & 6.9 & 47.7 & 66.5 & 51.9 & 29.7 & 50.8 & 59.4 \\
            
            \hline
             Ours: &  &    &  &  &  &  &\\
             VFNet & R-50 & 19.3 & 44.3/44.8 & 62.5/63.1 & 48.1/48.7 & 26.7/27.2 & 47.3/48.1 & 54.3/54.8 \\
             VFNet & R-101 & 15.6 & 46.0/46.7 & 64.2/64.9 & 50.0/50.8 & 27.5/28.4 & 49.4/50.2 & 56.9/57.6 \\
             VFNet & X-101-32x4d & 13.1 & 46.7/47.6 & 65.2/66.1 & 50.8/51.8 & 28.3/29.4 & 50.1/50.9 & 57.3/58.4 \\
             VFNet & X-101-64x4d & 9.2 & 47.4/48.5 & 65.8/67.0 & 51.5/52.6 & 29.5/30.1 & 50.7/51.7 & 58.1/59.7 \\
             VFNet & R2-101~\cite{Res2Net} & 13.0 & 48.4/49.3 & 66.9/67.6 & 52.6/53.5 & 30.3/30.5 & 52.0/53.1 & 59.2/60.5 \\
             VFNet & R-50-DCN & 16.3 & 47.3/48.0 & 65.6/66.4 & 51.4/52.3 & 28.4/29.0 & 50.3/51.2 & 59.4/60.4 \\
             VFNet & R-101-DCN & 12.6 & 48.4/49.2 & 66.7/67.5 & 52.6/53.7 & 28.9/29.7 & 51.7/52.6 & 61.0/62.4 \\
             VFNet & X-101-32x4d-DCN & 10.1 & 49.2/50.0 & 67.8/68.5 & 53.6/54.4 & 30.0/30.4 & 52.6/53.2 & 62.1/62.9 \\
             VFNet & X-101-64x4d-DCN & 6.7 & 49.9/50.8 & 68.5/69.3 & 54.3/55.3 & 30.7/31.6 & 53.1/54.2 & 62.8/64.4 \\
             VFNet & R2-101-DCN~\cite{Res2Net} & 10.3 & 50.4/51.3 & 68.9/69.7 & 54.7/55.8 & 31.2/31.9 & 53.7/54.7 & 63.3/64.4 \\
             
             VFNet-X-800 & R2-101-DCN~\cite{Res2Net} & 8.0 & 53.7 & 71.6 & 58.7 & 34.4 & 57.5 & \textbf{67.5} \\
             VFNet-X-1200 & R2-101-DCN~\cite{Res2Net} & 4.2 & \textbf{55.1} & \textbf{73.0} & \textbf{60.1} & \textbf{37.4} & \textbf{58.2} & 67.0 \\
            
            \hline
        \end{tabular}
    \end{center}
    \vspace{-5mm}
\caption{Performance (single-model single-scale) comparison with state-of-the-art detectors on MS COCO \texttt{test-dev}. VFNet consistently outperforms the strong baseline ATSS by $\sim$2.0 AP. Our best model VFNet-X-1200 reaches 55.1 AP, achieving the new stat-of-the-art. 'R': ResNet. 'X': ResNeXt. 'R2': Res2Net. 'DCN': Deformable convolution network. '/' separates results of the MSTrain image scale range 1333$\times$[640:800] / 1333$\times$[480:960]. FPSs with $^{\dag}$ are from papers.}
\label{table:sota}
\end{table*}
\begin{table}[t]
    \begin{center}
        \begin{tabular}{ K{12em}  | K{3em} K{3em} K{3em} }
            \hline
             Method  & AP &  AP$_{50}$ &  AP$_{75}$ \\
            \hline
             RetinaNet~\cite{retinaNet} + FL  & 36.5 & 55.5 & 38.8  \\ 
             RetinaNet~\cite{retinaNet} + GFL & 37.3 & 56.4 & 40.0  \\ 
             RetinaNet~\cite{retinaNet} + VFL & \textbf{37.4} & 56.5 & 40.2  \\ 
             \hline
             FoveaBox~\cite{foveaBox} + FL  & 36.3 & 56.3 & 38.3  \\ 
             FoveaBox~\cite{foveaBox} + GFL & 36.9 & 56.0 & 39.7  \\ 
             FoveaBox~\cite{foveaBox} + VFL & \textbf{37.2} & 56.2 & 39.8  \\
             \hline
             RepPoints~\cite{repPoints} + FL  & 38.3 & 59.2 & 41.1  \\ 
             RepPoints~\cite{repPoints} + GFL & 39.2 & 59.8 & 42.5 \\ 
             RepPoints~\cite{repPoints} + VFL & \textbf{39.7} & 59.8 & 43.1  \\ 
             \hline
             ATSS~\cite{ATSS} + FL & 39.3 & 57.5 & 42.5 \\
             ATSS~\cite{ATSS} + GFL & 39.8 & 57.7 & 43.2 \\
             ATSS~\cite{ATSS} + VFL & \textbf{40.2} & 58.2 & 44.0 \\
             \hline
             VFNet + FL  & 40.0 & 58.0 & 43.2 \\ 
             VFNet + GFL & 41.1 & 58.9 & 42.2 \\ 
             VFNet + VFL & \textbf{41.6} & 59.5 & 45.0 \\ 
            \hline
        \end{tabular}
    \end{center}
    \vspace{-5mm}
\caption{Comparison of performances when applying the focal loss (FL)~\cite{retinaNet}, the generalized focal loss (GFL)~\cite{GFL} and our varifocal loss (VFL) to existing popular dense object detectors and our VFNet.}
\label{table:existing}
\vspace{-4mm}
\end{table}

\vspace{0.3mm}
\subsection{Ablation Study}
\subsubsection{Varifocal Loss}
We first investigate the effect of the hyper-parameters of the varifocal loss on the detection performance. There are two hyper-parameters: \textbf{$\alpha$} for balancing the losses between positive examples and negative examples, and \textbf{$\gamma$} for down-weighting the losses of the easy negative examples. 
We show the performance of the VFNet in Table~\ref{table:VFL} when varying $\alpha$ from 0.5 to 1.5 and $\gamma$ from 1.0 to 3.0 (only the results obtained with optimal $\alpha$  are shown). It shows that similar results above 41.2 AP are achieved and our varifocal loss is quite robust to different sets of ($\alpha$, $\gamma$). Among those, \textbf{$\alpha$ = 0.75} and \textbf{$\gamma$ = 2.0} work best (41.6 AP), and we adopt these two values for all the following experiments.

We also investigate the effect of weighting the loss of the positive example with the training target q, termed as \textit{q weighting}. The fourth row in Table~\ref{table:VFL} shows the performance of the optimal set of ($\alpha$, $\gamma$) without q weighting and 0.4 AP drop is observed (41.2 AP v.s 41.6 AP). This confirms the positive effect of q weighting.

\vspace{-3mm}
\subsubsection{Individual Component Contribution}

We study the impact of the individual component of our method and results are shown in Table~\ref{table:ablation}. The first row shows the performance of the raw VFNet (FCOS+ATSS without centerness branch) trained with the focal loss and 39.0 AP is acquired. Replacing the focal loss with our varifocal loss, the performance is improved to 40.1 AP, which is 0.9 AP higher than the FCOS+ATSS. By adding the star-shaped representation and bounding box refinement modules, the performance is further boosted to 40.7 AP and 41.6 AP respectively. These results verify the effectiveness of the three modules in our VFNet.

\subsection{Comparison with State-of-the-Art}
We compare our VFNet with other detectors on the COCO \texttt{test-dev}. We select the ATSS~\cite{ATSS} as our baseline since it has similar performance to the FCOS+ATSS.

Table~\ref{table:sota} presents the results. Compared with the strong baseline ATSS, our VFNet achieves $\sim$2.0 AP gaps with different backbones, \eg 46.0 AP vs. 43.6 AP with the ResNet-101 backbone. This validates the contributions of our method. Compared to the concurrent work, the GFL~\cite{GFL} (whose MSTrain scale range is 1333x[480:800]), our VFNet is consistently better than it by a considerable margin. Meanwhile, our model trained with Res2Net-101-DCN~\cite{Res2Net} achieves a single-model single-scale AP of 51.3, surpassing almost all recent state-of-the-art detectors.

We also report the inference speed of VFNet in terms of frame per second (FPS) on a Nvidia V100 GPU. Since it is difficult to get the speed of all the listed detectors under exactly same settings, we only compare VFNet with the baseline ATSS. 
It can be seen that our VFNet is very efficient, \eg achieving 44.8 AP at 19.3 FPS, and only incurs small additional computation overhead compared to the baseline.

\vspace{-0.5mm}
\subsection{VarifocalNet-X}
To push the envelope of VFNet, We also implement some extensions to the original VFNet. This version of VFNet is called VFNet-X and those extensions include: 

\noindent \textbf{PAFPN.} We replace the FPN with the PAFPN~\cite{PAFPN}, and apply the DCN and group normalization (GN)~\cite{GN} in it.

\noindent \textbf{More and Wider Conv Layers.} We stack 4 convolution layers in the detection head, instead of 3 layers in the original VFNet, and increase the original 256 feature channels to 384 channels. 

\noindent \textbf{RandomCrop and Cutout.} We employ the random crop and cutout~\cite{cutout} as additional data augmentation methods.

\noindent \textbf{Wider MSTrain Scale Range and Longer Training.} We adopt a wider MSTrain scale range, from 750$\times$500 to 2100$\times$1400, and initially train the VFNet-X for 41 epochs.

\noindent \textbf{SWA.} We apply the technique of stochastic weight averaging (SWA)~\cite{SWA} in training the VFNet-X, which brings 1.2 AP gain. Specifically, after the initial 41-epoch training of VFNet-X, we further train it for another 18 epochs using a cyclic learning rate schedule and then simply average those 18 checkpoints as our final model.

\begin{figure*}[tbp]
	\centering
	\begin{tabular}{cc}
		\includegraphics[width=0.48\textwidth]{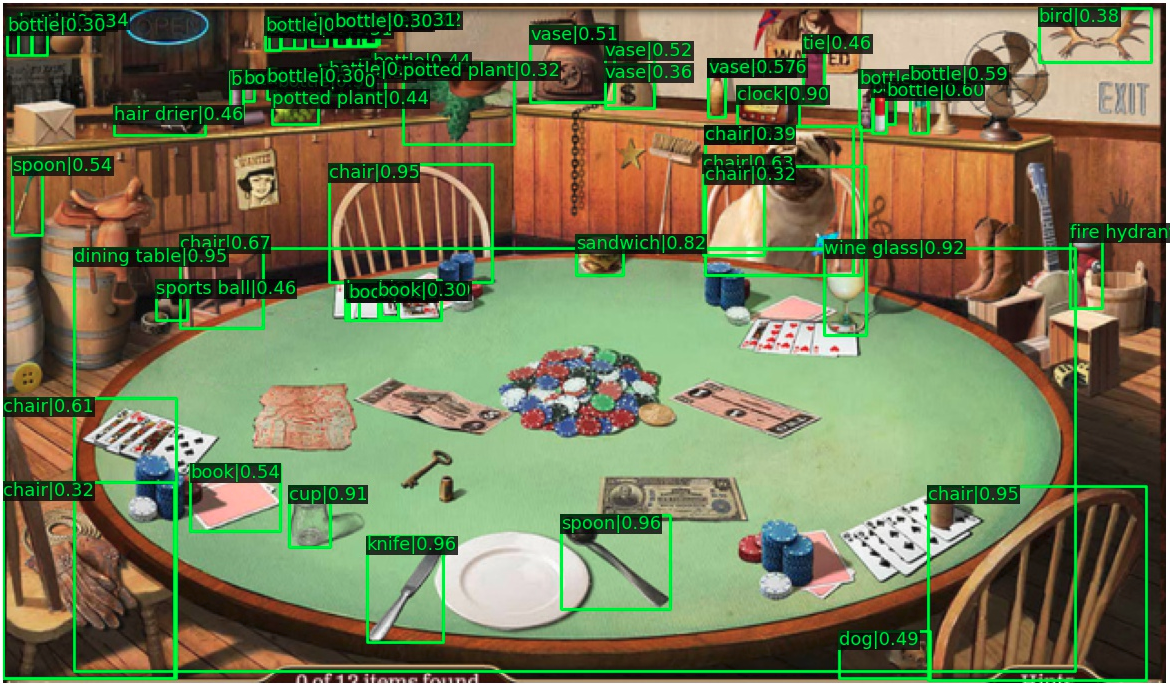}&
		\includegraphics[width=0.48\textwidth]{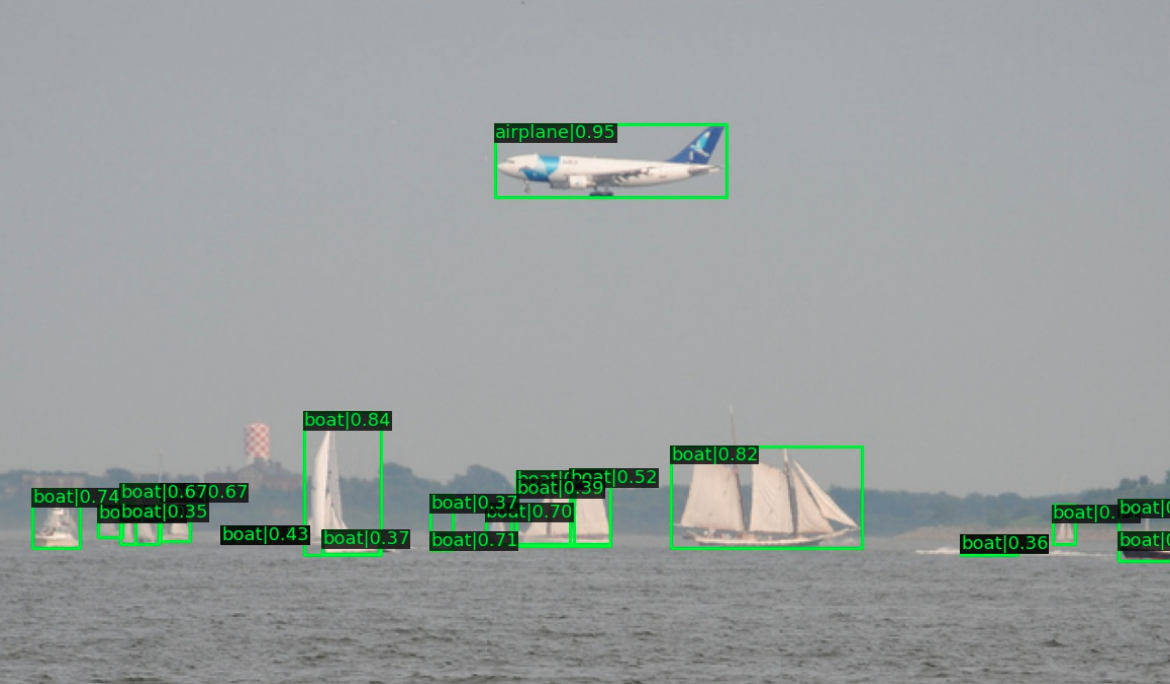}\\
		\includegraphics[width=0.48\textwidth]{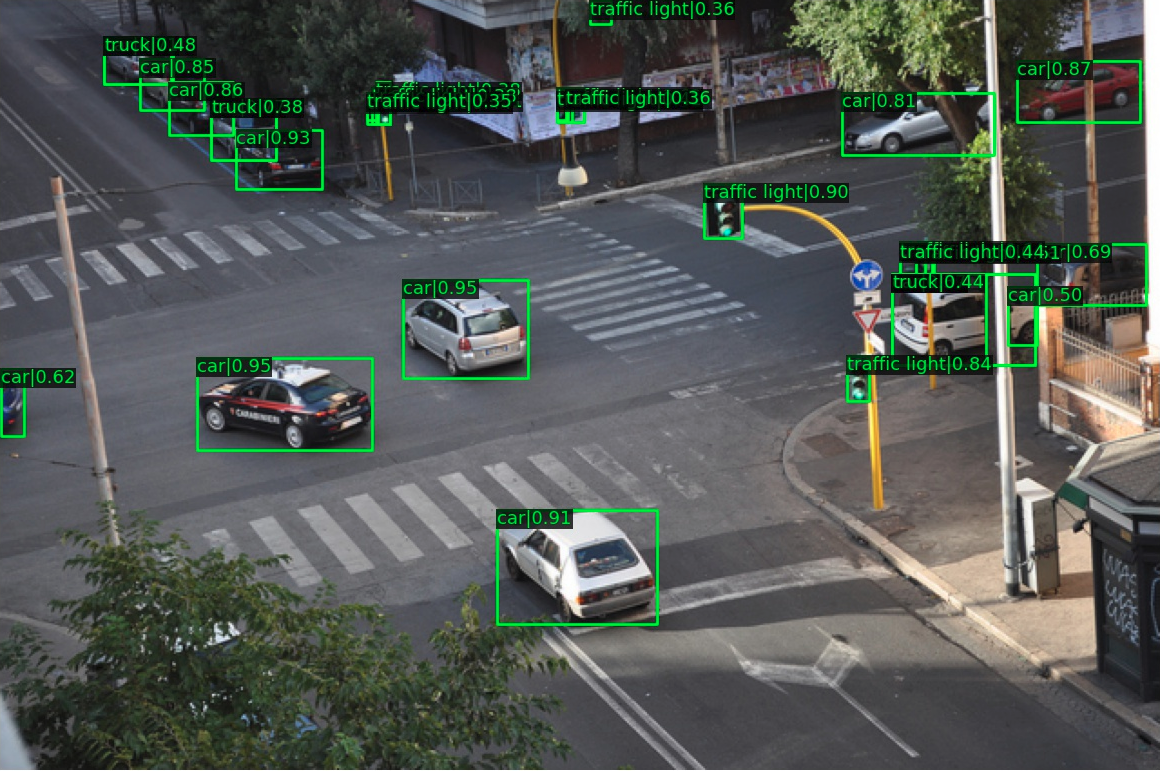}&
		\includegraphics[width=0.48\textwidth]{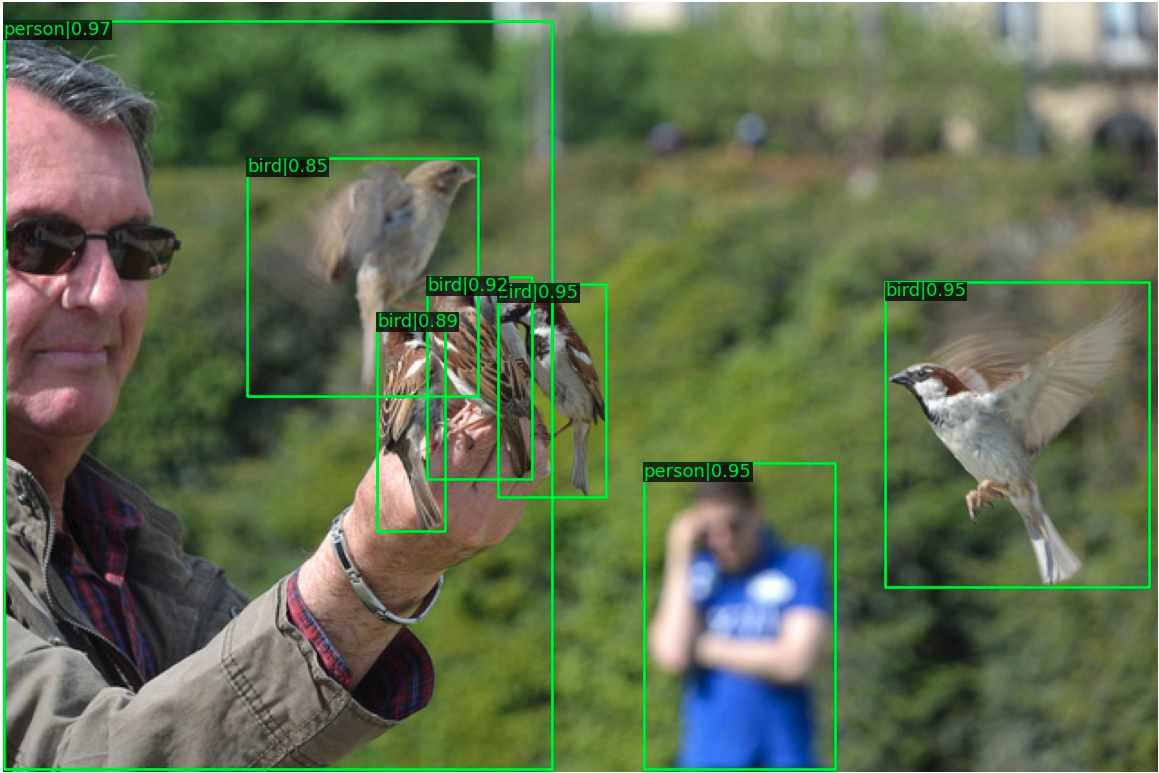}\\
		\includegraphics[width=0.48\textwidth]{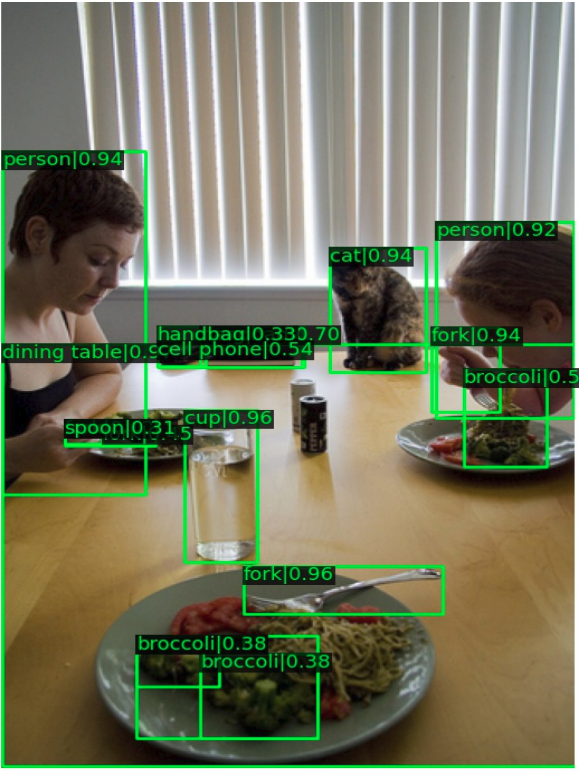}&
		\includegraphics[width=0.48\textwidth]{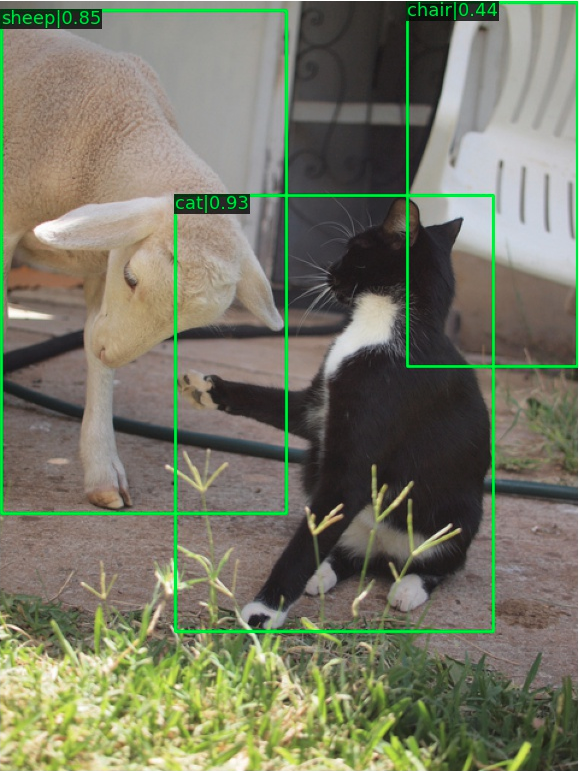}\\
	\end{tabular}
	\caption{Detection examples of applying our best model on COCO \texttt{test-dev}. The score threshold for visualization is 0.3.
	}\vspace{-2mm}
	\label{fig:example}
\end{figure*}

The performance of VFNet-X on COCO \texttt{test-dev} is shown in the last rows of Table~\ref{table:sota}. When the inference scale 1333$\times$800 and soft-NMS\cite{SoftNMS} are adopted, VFNet-X-800 achieves 53.7 AP, while simply increasing the image scale to 1800$\times$1200, VFNet-X-1200 reaches a new state-of-the-art \textbf{55.1} AP, surpassing prior detectors by a large margin. 
 Qualitative detection examples of applying this model to the COCO \texttt{test-dev} can be found in Figure~\ref{fig:example}.

\vspace{-0.5mm}
\subsection{Generality and Superiority of Varifocal Loss}
\vspace{-0.5mm}
To verify the generality of our varifocal loss, we apply it to some existing popular dense object detectors, including RetinaNet~\cite{retinaNet}, FoveaBox~\cite{foveaBox}, RepPoints~\cite{repPoints} and ATSS~\cite{ATSS}, and evaluate the performance on the \texttt{val2017}. We simply replace the focal loss (FL)~\cite{retinaNet} used in these detectors (ResNet-50 backbone) with our varifocal loss for training. We also train them with the generalized focal loss (GFL) for comparison. 

Table~\ref{table:existing} shows the results.
We can see that our varifocal loss improves RetinaNet, FoveaBox and ATSS consistently by 0.9 AP. For RepPoints, the gain increases to 1.4 AP. This shows that our varifocal loss can easily bring considerable performance boost to existing dense object detectors. Compared to the GFL, our varifocal loss performs better than it in all cases, evidencing the superiority of our varifocal loss. 

Additionally, we train our VFNet with the FL and GFL for further comparison. Results are shown in the last section of Table~\ref{table:existing} and the consistent advantage of our varifocal loss over the FL and GFL can be observed.

\vspace{-2mm}

\section{Conclusion}
\vspace{-1mm}
In this paper, we propose to learn the IACS for ranking detections. We first show the importance of producing the IACS to rank bounding boxes and then develop a dense object detector, VarifocalNet, to exploit the advantage of the IACS. In particular, we design a varifocal loss for training the detector to predict the IACS, and a star-shaped bounding box feature representation for IACS prediction and bounding box refinement. Experiments on the MS COCO benchmark verify the effectiveness of our methods and show that our VarifocalNet achieves the new stat-of-the-art performance among various object detectors.

{\small
\bibliographystyle{unsrt}
\bibliography{VarifocalNet}
}

\end{document}